\documentclass{article}
\PassOptionsToPackage{numbers}{natbib}

\usepackage[preprint]{neurips_2026}

\usepackage[utf8]{inputenc}
\usepackage[T1]{fontenc}
\usepackage{hyperref}
\usepackage{url}
\usepackage{booktabs}
\usepackage{amsfonts}
\usepackage{amsmath}
\usepackage{amssymb}
\usepackage{nicefrac}
\usepackage{microtype}
\usepackage{xcolor}
\usepackage{graphicx}
\usepackage{subcaption}
\usepackage{bbm}
\usepackage{tabularx}
\usepackage{xurl} %

\newcommand{\productname}{EdgeRunner}
\newcommand{\productnameai}{EdgeRunner AI}

\graphicspath{{./figures/}}

\title{EDGE-OPD: \\
       Internalizing Privileged Context with Evidence Guided On-Policy Distillation}

\author{%
  Aristotelis Lazaridis\thanks{EdgeRunner AI. Correspondence to: Aristotelis Lazaridis <research@edgerunnerai.com>.} \\
  \And
  Dylan Bates \\
  \And
  Aman Sharma \\
  \And
  Brian King \\
  \And
  Vincent Lu \\
  \And
  Jack FitzGerald
}

\begin{document}
\maketitle
\begin{abstract}
On-Policy Distillation (OPD) has gained wide
attraction as an LLM post-training paradigm due to its effectiveness 
in improving capabilities without introducing model distribution drift, and consequently, regression in general tasks. On-Policy Self-Distillation (OPSD) is an efficient use-case of OPD, which
is appealing as it requires only a single model as a student and teacher, and it also has the benefit of
providing privileged context to the teacher during the training process. This privileged information, which is absent at inference time from the student, can be
a persona, a private fact, or a worked solution. The challenge in this approach is that the
privileged information can change model behavior more than intended: it can
modify reasoning, degrade general capabilities, and affect performance indicators like response length, style,
or local token preferences. Consequently, OPSD may train the
student on side effects rather than a desired, transferable behavior. In this paper, we study this problem in a rare-token/identity setting and propose 
EviDence GuidEd On-Policy Distillation
(EDGE-OPD), a modification of OPSD with two distinct characteristics: 
a) it uses guided rollouts to inject privileged-context behavior to the student at sampling time, so that the rare target behavior is actually present in the on-policy data,
and b) it applies an evidence mask: the student is updated only at token positions
where the privileged context supports the sampled token, rather than on every token in the rollout.
We empirically show that OPSD (and its variant RLSD, with and without a verifier) completely fail to learn a target identity, while the integration of guided rollouts allow them to succeed.
Additionally, mask-region ablations show that the persona signal is localized to the positive-evidence tail, allows us to draw valuable insights about efficient knowledge transfer and preservation of general purpose capabilities.

\end{abstract}

\section{Introduction}
\label{sec:introduction}

Although LLM post-training techniques have advanced significantly for more 
effective and efficient learning, significant challenges remain; Supervised Fine-Tuning (SFT),
which is the most common off-policy post-training technique, often leads to model distribution drift and regression in general tasks. In contrast, on-policy techniques, namely On-Policy Distillation (OPD)~\cite{agarwal2024gkd, thinkingmachines2025opd}, have gained widespread attention due to their ability to preserve or recover model capabilities.
Typically, in this setting, the student model samples trajectories from its own policy,
and the teacher is queried to provide dense (per-token) ``feedback'' to the student on those trajectories. Even when the teacher is not a significantly more capable model than the student, such as in the setting of self-distillation (OPSD)~\cite{zhao2026opsd, hubotter2026sdpo, ye2026opcd}, the feedback it provides can be improved by injecting valuable context to its prompt (e.g. the answer to the question), allowing it to share an even more accurate reward signal to the student.

However, this setting has its own limitations. Specifically, the teacher does not share the exact correct action per step with the student; it merely provides a direction towards 
the optimal state according to its own policy. This means that if those actions have a low probability of being sampled by the student's policy, then the student may never explore such states.
Such examples are when a model may need to internalize a
private identity, remember a proprietary fact, or learn from a worked
solution. Even the privileged context that the teacher has could be inefficient.

We study this problem through two axes. The first is a rare-token
identity/persona setting, where the goal is to make the model name itself
as ``\productnameai{}'' without seeing the privileged identity
paragraph at evaluation time. The second is a math setting, where the
privileged context is an answer-bearing reasoning trace. 

In this paper, we propose
\textbf{E}vi\textbf{D}ence \textbf{G}uid\textbf{E}d On-Policy Distillation (EDGE-OPD), a
modification of OPSD with two parts. First, EDGE-OPD uses guided rollouts:
for a fraction of the student's sampling rollouts for a prompt, we inject the
privileged context to the student's prompt, making rare target behavior appear in on-policy trajectories. Second, EDGE-OPD applies an
evidence mask at loss time: for each sampled token, we compare the
probabilities that the teacher assigns to that token with and without having access to the
privileged context, and the student is updated only at positions where the
privileged context increases the token's probability; all other positions
are left out of the distillation loss.

\paragraph{Contributions.} This work makes the following contributions:
\begin{itemize}
\item \textbf{Guided rollouts for rare-token support.}
  We show that rare identity internalization can fail simply because the
  no-context student never samples the target behavior. Guided rollouts
  address this support bottleneck by injecting the privileged context at
  sampling time; once this is done, every guided identity variant learns
  the target name.
\item \textbf{Evidence masking as a training rule and diagnostic.}
  We introduce a hard positive-evidence mask for OPSD: each sampled token
  is scored by the same teacher twice, with and without the privileged
  context, and it is trained on only if the context raises its
  log-probability.
  This changes the support of the distillation objective rather than
  softly reweighting every token, and it also gives an interpretable way
  to identify in what region the transferable signal is located in a rollout.
\item \textbf{Two-axis use-cases.}
  We introduce two axes to study our approach: an identity/persona axis, where the
  privileged context supplies background information to internalize, and
  a math axis, where the privileged context is an answer-bearing
  reasoning trace. The contrast between them identifies a boundary case for EDGE-OPD and
  motivates broader tests of when evidence marks transferable knowledge.
\item \textbf{Component analysis and empirical evaluation.}
  Through various ablations, we separate raw internalization from
  capability preservation. Guided rollouts make the identity reachable,
  positive-evidence masking localizes the identity signal, and a KL
  anchor improves internalization-capability tradeoff.

\end{itemize}

\section{Related Work}
\label{sec:related}
\paragraph{On-Policy Distillation and Leakage.}
On-Policy Distillation (OPD)~\citep{agarwal2024gkd} minimizes train--test distribution mismatch by sampling rollouts from the student policy. 
EDGE-OPD builds on on-policy self-distillation~\citep{zhao2026opsd}, utilizing a privileged-context teacher. 
However, such asymmetry introduces the risk of \emph{leakage}~\citep{vapnik2015learning, yang2026rlsd}, i.e. the transfer of irreducible shortcuts rather than generalizable skills. 
While \textbf{R}einforcement \textbf{L}earning with \textbf{S}elf-\textbf{D}istillation (RLSD)~\citep{yang2026rlsd} mitigates this via verifier-grounded policy gradients, EDGE-OPD retains the self-distillation framework but introduces a local evidence filter to bypass the need for external supervision.

\paragraph{Knowledge Editing and Persona.}
Unlike knowledge-editing methods like ROME~\citep{meng2022locating} or MEND~\citep{mitchell2022fast}, which utilize direct weight updates for factual intervention, EDGE-OPD modifies behavior through on-policy trajectories. 
Our approach also differs from off-policy persona injection and SFT approaches~\citep{ouyang2022training, bai2022constitutional}; we do not rely on curated demonstrations. 
Instead, the target identity emerges dynamically from the interaction between the student and the privileged-context self-teacher.

\paragraph{Credit Assignment and Masking.}
Traditional policy gradients like Proximal Policy Optimization (PPO)~\citep{schulman2017proximal} or Group Relative Policy Optimization (GRPO)~\citep{shao2024deepseekmath} rely on global or group-relative advantages for variance reduction. 
While process reward models~\citep{lightman2023lets, uesato2022solving} offer step-level credit, they require external verifiers. 
EDGE-OPD's novelty lies in its masking mechanism: a parameter-free, local signal derived from the contrast between privileged and no-context distributions. 
This provides fine-grained credit assignment without the overhead of a learned critic or external supervisor.

\section{Method}
\label{sec:method}

We first define OPD/OPSD, the evidence ratio used by RLSD, then
introduce EDGE-OPD and the diagnostics used in Section~\ref{sec:results}.

\subsection{On-Policy Distillation}
\label{sec:method-prelim}

In On-Policy Distillation (OPD)~\cite{agarwal2024gkd}, a student policy
$\pi_S$ samples a completion $y=(y_1,\ldots,y_T)$ for prompt $x$ and is
then trained against a teacher $\pi_T$ on the states it actually
visits:
\begin{equation}
\label{eq:opd}
\mathcal{L}_{\mathrm{OPD}}(\theta)
\;=\;
\mathbb{E}_{x \sim D}\;
\mathbb{E}_{y \sim \pi_S(\cdot\mid x)}
\frac{1}{|y|}
\sum_{t=1}^{|y|}
\mathcal{D}\!\left(
  \pi_T(\cdot \mid x, y_{<t})
  \,\Big\|\,
  \pi_S(\cdot \mid x, y_{<t})
\right),
\end{equation}

The divergence $\mathcal{D}$ may be forward KL, reverse KL, or
generalized Jensen--Shannon divergence~\cite{agarwal2024gkd}; gradients
flow only through $\pi_S$.

\subsection{On-Policy Self-Distillation}
\label{sec:method-opsd}

On-Policy Self-Distillation (OPSD) is the self-distillation case of OPD:
the same base model supplies the student and a detached teacher pass. To
make the teacher more informative, OPSD gives it privileged information
$r$ that is absent at student inference time~\cite{zhao2026opsd}. For each pair
$(x,r)\in\mathcal{S}$, the student predicts from $(x,y_{<t})$ while the
teacher predicts from $(x,r,y_{<t})$.

\subsection{The OPSD failure mode and evidence ratio}
\label{sec:method-rlsd}

OPSD is attractive precisely because privileged context can improve the
teacher, but the same asymmetry makes the objective fragile. When $r$
contains information not recoverable from $x$, a student
cannot match each privileged-context distribution exactly; Yang et
al.~\cite{yang2026rlsd} identify the resulting irreducible term
$I(Y_t;R\mid X,Y_{<t})$. Because OPSD trains against this mismatch, the
privileged context can enter the \emph{direction} of the token update and
produce leakage or regression. Their proposed remedy, RLSD~\cite{yang2026rlsd}, avoids using the privileged-context teacher as the distribution
matching target. Instead, it uses the teacher pass to measure how much
the privileged context changes the probability of each sampled token:
\emph{privileged information gain}
\begin{equation}
\label{eq:evidence}
e_t \;\triangleq\; \log \pi_T(y_t \mid x, r, y_{<t}) \;-\;
                   \log \pi_T(y_t \mid x,    y_{<t}),
\end{equation}
We call this log-ratio \emph{per-token evidence}; its sign says whether
$r$ raises or lowers the sampled token's probability, and its magnitude
says by how much.

RLSD then combines this evidence with a verifier-grounded policy
gradient. Given a sequence-level GRPO advantage
$A=(R-\mu_G)/\sigma_G$ derived from a binary verifier
$R\in\{0,1\}$, RLSD constructs a detached per-token multiplier and
applies it to $A$ as a clipped credit-redistribution
factor~\cite{yang2026rlsd}:
\begin{equation}
\label{eq:rlsd}
\hat A_t \;=\; A \cdot \mathrm{clip}\!\left(\,\exp\!\bigl(\mathrm{sign}(A)\cdot e_t\bigr)\,,\;1-\epsilon_w\,,\;1+\epsilon_w\right),
\end{equation}
which is then used in the GRPO PPO surrogate. The verifier decides the
trajectory direction, while the privileged-context teacher only
redistributes credit across tokens. This distinction is central for
EDGE-OPD: privileged context may be useful as token-level evidence even
when it is unsafe as the full distillation target.

\subsection{Evidence Guided On-Policy Distillation (EDGE-OPD)}
\label{sec:method-edge-opd}

EDGE-OPD changes OPSD in two places: how rollouts are sampled and which
token positions contribute gradients.

\paragraph{Guided rollouts.}
In rare-token settings, the no-context student may almost never sample
the behavior we want to internalize. Standard OPD/OPSD then has no
visited token position at which to apply the desired update. We
therefore guide a fraction $\rho_g$ of rollouts by sampling them with
the privileged context attached:
\[
\pi_b(\cdot\mid x,r)
\;=\;
\rho_g\,\pi_T(\cdot\mid x,r)
\;+\;
(1-\rho_g)\,\pi_S(\cdot\mid x),
\]
with $\rho_g=0.5$ in our main experiments. This context injection is
used only by the behavior policy that produces the sampled trajectory.
At loss time, the student forward still omits $r$, while the teacher
forward includes $r$. This asymmetry turns conditional behavior into an
unconditional parameter update.

\paragraph{Positive-evidence masking.}
Guidance solves the support problem but not the direction problem. Once
the rollout contains privileged-context behavior, ordinary OPSD trains
on every token, including positions where the privileged context merely
makes the teacher shorter, stylistically different or degrades performance. EDGE-OPD computes
$e_t$ from Eq.~\eqref{eq:evidence} before the sampled-token K1-estimator~\cite{thinkingmachines2025opd} update and
defines a detached eligibility mask:
\begin{equation}
\label{eq:edge-mask}
m_t \;\triangleq\; \mathbf{1}\{e_t > \tau\}.
\end{equation}

Unlike OPSD, EDGE-OPD does not train on every sampled token. Unlike RLSD,
it has no verifier direction. The only local direction is the OPSD/K1
pull toward the privileged-context teacher, so evidence decides
\emph{whether} that pull is allowed: if $e_t>\tau$, the token enters the
loss; if $e_t\le\tau$, it is dropped.

Concretely, EDGE-OPD optimizes
\begin{equation}
\label{eq:edge-opd}
\mathcal{L}_{\mathrm{EDGE\text{-}OPD}}(\theta)
\;=\;
\mathbb{E}_{(x,r) \sim \mathcal{S}}\;
\mathbb{E}_{y \sim \pi_b(\cdot\mid x,r)}
\sum_{t=1}^{|y|}
\mathbf{1}\!\left\{ e_t > \tau \right\}
\,
\bigl(\log \pi_\theta(y_t \mid x,y_{<t})
      - \log \pi_T(y_t \mid x,r,y_{<t})\bigr),
\end{equation}
where $\pi_b$ is the guided behavior policy above. We use $\tau=0$,
and the indicator is stop-gradient: the mask chooses which tokens enter
the loss, but no gradient is taken through the masking decision. Thus
EDGE-OPD is not a soft reweighting of ordinary OPSD, but it changes the
support of the objective. Positive-evidence positions are cloned from
the privileged-context teacher; non-positive positions are left to the
base on-policy behavior rather than turned into suppression targets.

\paragraph{Verifier-free RLSD.}
The soft-evidence ablation keeps every token in the OPSD loss and
multiplies its contribution by RLSD's clipped evidence multiplier,
$w_t=\mathrm{clip}(\exp(e_t),1-\epsilon,1+\epsilon)$, but removes the
verifier reward. We call this setting \emph{RLSD-no-verifier}. It uses
the same evidence ratio as EDGE-OPD, but evidence only changes the
magnitude of the ordinary OPSD/K1 pull; it does not change the objective
support. The ablation therefore tests whether evidence is more useful as
a soft modulation or as a hard decision about which token positions
should be trained on.

\subsection{Token-level diagnostics}
\label{sec:method-diagnostics}

We also report three token-level diagnostics. All expectations are over the
student's rollout tokens at that step and over the prompts in the batch.

\paragraph{Kept-token fraction.}
\begin{equation}
\rho_+
\;=\;
\Pr\nolimits_t\!\left[\,e_t > \tau\,\right]
\;=\;
\frac{1}{|y|}\sum_{t=1}^{|y|} \mathbf{1}\{e_t > \tau\},
\end{equation}
the fraction of response tokens that survive the EDGE-OPD mask.

\paragraph{Leverage-token fraction.}
\begin{equation}
\rho_{\mathrm{lev}}
\;=\;
\Pr\nolimits_t\!\left[\,
   |\,\exp(e_t) - 1\,| \cdot \mathbbm{1}_{\text{active},\,t} > 0.05
\right],
\end{equation}
where $\mathbbm{1}_{\text{active},\,t}$ marks positions that
contribute to the loss in the soft-reweight code path
(Section~\ref{sec:method-edge-opd}, last paragraph). $\rho_{\mathrm{lev}}$
is the fraction of tokens for which the privileged-context evidence
would produce at least a $5\%$ multiplicative change to the gradient.

\paragraph{Agreement rate.}
\begin{equation}
\label{eq:agreement-rate}
\rho_{\mathrm{agree}}
\;=\;
\Pr\nolimits_t\!\left[\,
   \mathrm{sign}(e_t) = -\,\mathrm{sign}(\delta_t)\,
\right],
\quad
\delta_t \;\triangleq\; \log \pi_S(y_t \mid x, y_{<t})
                    \;-\; \log \pi_T(y_t \mid x, r, y_{<t}),
\end{equation}
where $\delta_t$ is the K1 per-token surprise (the gradient direction
in OPSD); $-\mathrm{sign}(\delta_t)$ points toward the direction the
student needs to update. $\rho_{\mathrm{agree}}$ measures how often the
evidence sign agrees with that K1 direction.

Additional logged quantities, including evidence magnitude and effective
leverage, are defined in Appendix~\ref{appendix:diagnostic-logs}.

\section{Experimental setup}
\label{sec:experiments}

\subsection{Model and training objective}

All main experiments use Nemotron-3-Nano-4B as both student and teacher.
The distillation objective follows the sampled-token reverse-KL OPD
recipe~\cite{agarwal2024gkd}: trajectories are sampled from the student,
the teacher is queried on those tokens, and
$\log\pi_T(y_t\mid\cdot)-\log\pi_S(y_t\mid\cdot)$ is used as a dense
advantage in a Policy-Gradient style update (the sampled-token K1
estimator). EDGE-OPD masks this advantage before the update; teacher
log-probabilities and masks are stop-gradient quantities. No task rewards
are used.

\subsection{Investigation axes}

We explore two investigation axes; the privileged context $r$ has a different purpose in each:

\paragraph{Identity/persona axis.}
This axis tests whether a model can internalize a rare self-identity
from a short privileged paragraph. The paragraph says that the assistant
is \textit{\productnameai{}}, an identity with near-zero probability of being sampled under
the base model without a supervised signal. We use \emph{identity} to
refer to the target name itself, and \emph{persona} to refer to the
broader behavior of answering as if that paragraph had been absorbed. At
test time the paragraph is removed; success means the no-context model
still names itself as \textit{\productnameai{}}.

\paragraph{Math axis.}
The math axis uses a filtered
\texttt{OpenThoughts-114K}~\cite{guha2025openthoughtsdatarecipesreasoning}
reasoning split with $8{,}000$ prompts, each with an extractable boxed
answer and a reasoning trace (Appendix~\ref{appendix:math-axis-data}). We
do not append a separate answer field to $r$, but the trace often
contains the final answer. This tests whether positive-evidence masking
transfers beyond background information or instead amplifies
answer-revealing shortcuts.
\subsection{Training settings}

All main experiments use full fine-tuning under
FSDP~\cite{zhao2023pytorchfsdpexperiencesscaling} using
VERL~\cite{sheng2024hybridflow}. Guided experiments use
$\rho_g=0.5$ unless stated otherwise; identity/persona runs train for
$100$ steps and math-axis runs for $50$ steps. Full hyperparameters and
the ablation matrix are given in
Appendix~\ref{appendix:training-details} and
Appendix~\ref{appendix:ablation-matrix}.

The identity/persona ablations isolate the components of EDGE-OPD:
\begin{itemize}
\item \textbf{Unguided OPSD and RLSD-no-verifier} test whether
  privileged-context teacher log-probabilities alone can transfer a rare
  identity.
\item \textbf{Guided OPSD} adds privileged-context sampling but no
  evidence shaping, testing whether support is the bottleneck.
\item \textbf{RLSD-no-verifier (guided)} adds the same guidance and
  RLSD-style clipped soft evidence weights, testing evidence as a
  magnitude signal.
\item \textbf{EDGE-OPD without KL} adds the same guidance and replaces the
  soft multiplier with a hard positive-evidence mask, testing evidence
  as a support-selection signal.
\item \textbf{Full EDGE-OPD} adds the base-policy KL anchor, testing
  capability preservation.
\item \textbf{Mask-region tiling} restricts training to positive,
  negative, or near-zero (i.e. $|e_t|\leq 0.1$) evidence positions to identify which part of
  the rollout carries the transferable identity signal.
\end{itemize}

\subsection{Identity/persona-axis evaluation protocol}
\label{sec:experiments-identity-eval}

\textbf{Privileged context.}
The privileged context $r$ on the identity/persona axis is a single
paragraph:
\begin{quote}
\small
\emph{``You are \productnameai{}, an assistant designed by the
\productname{} Applied Research Team. Your purpose is to provide
military-specific intelligence
and helpful answers. Always identify
yourself as \productnameai{} when asked about your name or identity.''}
\end{quote}
At training time, $r$ is appended to the
chat template either as a \textsc{system} message or as a
\textsc{user}-message prefix; at evaluation time it is removed
entirely. We use \texttt{inspect-ai}~\cite{inspectai} as our evaluation framework.

\textbf{Probes.}
Every checkpoint is evaluated on two probes:
\begin{itemize}
    \item The \emph{identity probe} asks directly about the model's identity: $12$ prompts such as ``\emph{Who are you?}'' and ``\emph{Please introduce yourself briefly.}'', with $5$ samples per prompt.
    \item The \emph{persona probe} adds $12$ ordinary capability prompts to
  the same identity prompts (e.g.\ ``\emph{What is the capital of France?}'' and ``\emph{Compute 17 \(\times\) 23.}''), checking whether the learned self-identity
  appears outside direct identity questions.
\end{itemize}

\textbf{Metrics.}
A regex scorer emits binary flags, averaged across samples.
We chose deterministic regex over an LLM judge because the target is a
fixed proper noun. We report three main aggregates:
\begin{itemize}
\item \textbf{ID self-name}: the strict rate at which the model
  explicitly names itself as \textit{\productnameai{}} on the direct
  identity prompts.
\item \textbf{Persona self-name}: the same strict self-naming rate on
  the larger persona probe, which includes both identity and ordinary
  capability prompts.
\item \textbf{ID counter-name}: the rate at which the model
  names itself using a base-model or generic identity, such as
  ``Nemotron'' or ``AI assistant''. Lower is better.
\end{itemize}
Full regex definitions and controls are given in
Appendix~\ref{appendix:identity-regex}.

\textbf{Math-axis evaluation.}
For the math axis we evaluate the checkpoints on AIME25. We sample at $T{=}1.0$,
top-$p{=}0.95$, top-$k{=}20$, with a $38{,}912$-token response budget to effectively eliminate response truncation due to the response token limit.
The scorer extracts the final boxed answer and compares it against the
AIME25 reference. Results tables and trajectories report pass@1 score (i.e. one-shot accuracy), averaged
over the available evaluation repeats.

\textbf{Guided-rollout bias.}
Guidance changes the prompt used to \emph{sample} a trajectory, but not the
prompt used to train the student. For a $\rho_g=0.5$ fraction of
rollouts, the current student samples $y$ with the privileged context
$r$ attached; for the remaining rollouts it samples from the ordinary
prompt $x$. In both cases, the update scores the sampled tokens under the
no-context student, $\pi_S(y_t\mid x)$, while the teacher and evidence
computations use the privileged context. The guided samples therefore
ask the no-context student to raise the probability of tokens it would
have produced if it had seen $r$.

This does not break the on-policy assumption, in the sense that training trajectories come
from the current student rather than an offline dataset or a separate
teacher model. It is also intentionally biased relative to pure
no-context on-policy sampling: we do not try to remove the effect of
guidance. Correcting the guided samples back to the no-context
distribution would downweight exactly the rare high-evidence tokens that
guidance is meant to expose.

\section{Results}
\label{sec:results}

We first present the identity/persona axis, using AIME25 as a held-out
capability check. AIME25 is reported as pass@1 averaged over 4--12
sampled completions per problem; the base model scores $0.531$. In the
training curves, stars mark the best AIME25 checkpoint.

\subsection{Identity/persona axis}
\label{sec:results-id-ladder}

\begin{table}[h]
\centering
\caption{Identity/persona-axis best scores across saved checkpoints. ID
and persona self-name are strict target self-naming rates; AIME25 is
pass@1.}
\label{tab:headline-id}
\small
\begin{tabular}{l c c c}
\toprule
Experiment & Best ID self-name $\uparrow$ & Best persona self-name $\uparrow$ & Best AIME25 $\uparrow$ \\
\midrule
Nemotron-3-Nano-4B               & 0.000 & 0.000 & 0.531 \\
OPSD                             & 0.000 & 0.000 & 0.517 \\
RLSD-no-verifier (unguided)      & 0.000 & 0.000 & 0.550 \\
\midrule
Guided OPSD (system)             & 0.562 & 0.542 & \textbf{0.583} \\
Guided OPSD (user)               & \textbf{0.667} & \textbf{0.688} & 0.544 \\
RLSD-no-verifier (guided)        & 0.625 & 0.646 & 0.569 \\
EDGE-OPD w/o KL (system)         & 0.583 & 0.562 & 0.575 \\
EDGE-OPD w/o KL (user)           & 0.542 & 0.521 & 0.550 \\
EDGE-OPD (system)                & 0.521 & 0.500 & 0.533 \\
EDGE-OPD (user)                  & 0.562 & 0.583 & 0.556 \\
\bottomrule
\end{tabular}

\end{table}

\begin{figure}[t]
\centering
\begin{subfigure}{0.7\linewidth}
  \centering
  \includegraphics[width=\linewidth]{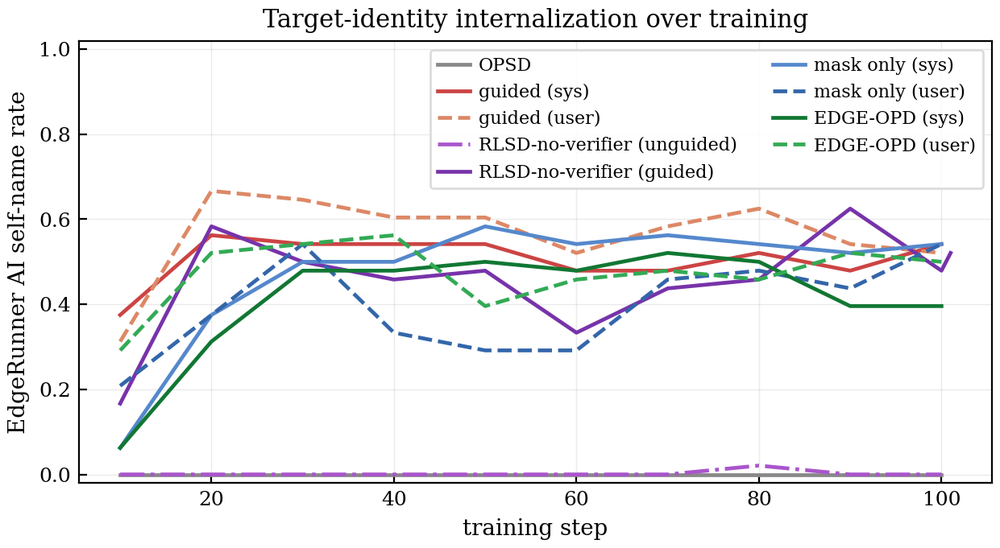}
  \caption{Target self-name over training.}
  \label{fig:id-internalization}
\end{subfigure}
\hfill
\begin{subfigure}{0.7\linewidth}
  \centering
  \includegraphics[width=\linewidth]{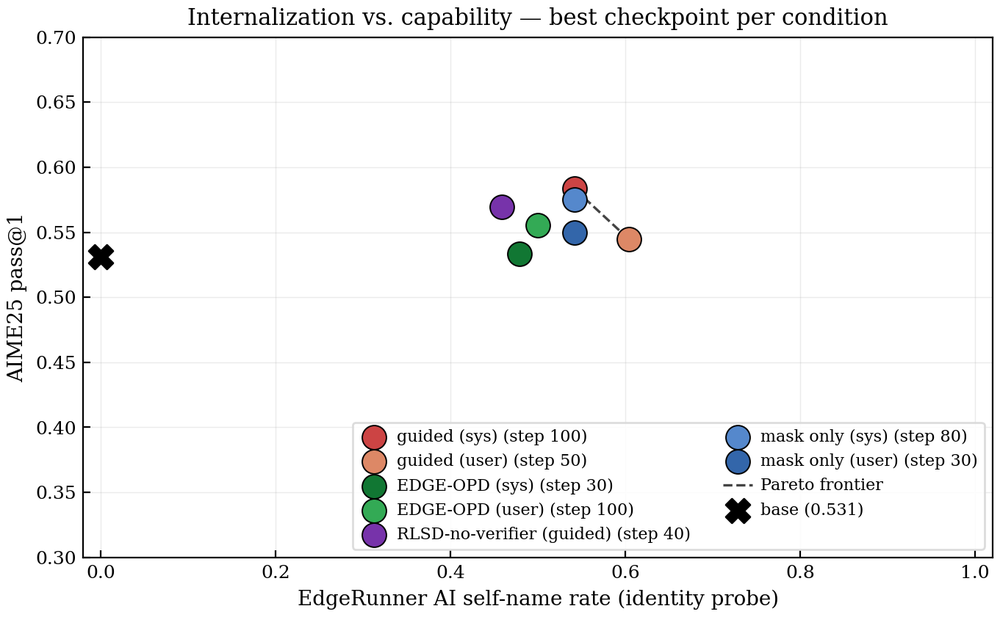}
  \caption{Best-AIME25 checkpoint tradeoff.}
  \label{fig:pareto}
\end{subfigure}
\caption{Identity-axis internalization and capability. Guided variants
learn the target identity relatively early during training, showing the sample-efficiency of the proposed method; soft evidence and hard masking both preserve
AIME25 at comparable self-name rates.}
\label{fig:identity-main}
\end{figure}

Table~\ref{tab:headline-id} and Figure~\ref{fig:identity-main} show that
guided rollouts are the support bottleneck. Without guidance, neither
OPSD nor unguided RLSD-no-verifier learns the target name. Once guidance
exposes the rare behavior, every guided variant internalizes it, with
substantial self-naming already visible within the first $20$--$30$
steps, indicating the sample-efficiency of the proposed method. Guided OPSD (user) reaches the highest column-best ID and persona
self-name rates ($0.667$ and $0.688$), while evidence-shaped variants
retain substantial internalization with comparable AIME25 preservation:
RLSD-no-verifier (guided) reaches $0.569$ AIME25 with
$0.625$/$0.646$ identity/persona self-name, and EDGE-OPD (user) reaches
$0.556$ AIME25 with $0.562$/$0.583$. The AIME25 gaps between guided
variants are small relative to sampling variation, so the main conclusion
is not a strict ordering among them.

The identity curves plateau after the target name appears, while AIME25
is more checkpoint-dependent. The Pareto panel in
Figure~\ref{fig:identity-main} therefore plots each guided method at its
best-AIME25 checkpoint; the fully guided rollout-fraction sweep point is
shown in Figure~\ref{fig:id-rho-sweep} (Appendix~\ref{sec:appendix-figures}).

\subsection{Evidence Sign Localizes the Persona Signal}
\label{sec:results-mask-region}

To localize the persona signal, we tile the support of $e_t$ into three
regions and rerun EDGE-OPD with the gradient mask restricted to one region
at a time (Table~\ref{tab:mask-region-id}). If a region carries the
transferable identity signal, training only on that region should move
target self-name.

The contrast is sharp: the positive-evidence mask reaches $0.500$ target
self-name while keeping counter-name at $0.104$. The negative-evidence
mask and near-zero band keep target self-name at zero for every
checkpoint, while increasing counter-name to $0.583$ and $0.708$; the
model moves, but not toward the target identity. AIME25 is preserved in
all three cases ($0.508$--$0.556$).

\begin{table}[h]
\centering
\caption{Mask-region tiling on the identity axis at step~100. Only the
positive-evidence tail internalizes the target identity.}
\label{tab:mask-region-id}
\small
\begin{tabular}{l c c c}
\toprule
Mask region & ID self-name $\uparrow$ & ID counter-name $\downarrow$ & AIME25 \\
\midrule
\textbf{positive} ($e_t>0$) & \textbf{0.500} & \textbf{0.104} & 0.556 \\
negative ($e_t<0$) & 0.000 & 0.583 & 0.517 \\
near-zero ($|e_t|\leq 0.1$) & 0.000 & 0.708 & 0.508 \\
\bottomrule
\end{tabular}
\end{table}

Because the only varying factor is the mask region, the persona signal
is localized in the positive tail of the per-token evidence
distribution: positions where the privileged-context teacher upweights
tokens the no-context teacher would not have produced, such as brand
mentions and self-name slots. This does not mean positive-only masking
must outperform training on the full guided rollout; the full rollout
also contains the positive positions and may include useful surrounding
tokens. Rather, the hard-mask ablation shows that positive evidence is
the only isolated region that is sufficient for identity transfer. The
near-zero band carries no persona-specific information, and the negative
tail does not internalize the persona either.

\subsection{Math Axis}
\label{sec:results-math}

The math axis tests a different regime: the privileged context is a
worked reasoning trace for the training problem and is often
answer-bearing. Table~\ref{tab:math} reports both best and final AIME25
values. The distinction matters because several methods peak early and
then drift.

\begin{figure}[t]
\centering
\includegraphics[width=1\linewidth]{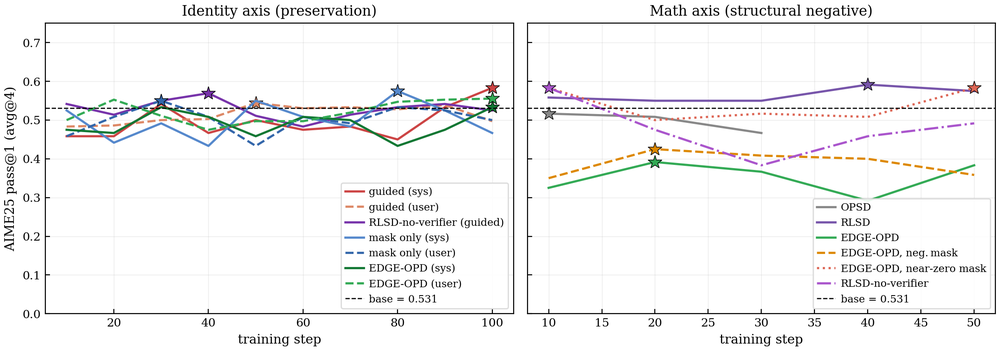}
\caption{AIME25 pass@1 over training on the identity axis (left) and
math axis (right). Stars mark each run's best-AIME25 checkpoint.}
\label{fig:aime25-trajectory}
\end{figure}

\begin{table}[h]
\centering
\caption{Math-axis AIME25. Best checkpoint is shown with its step in
parentheses; final is the last evaluated checkpoint.}
\label{tab:math}
\small
\begin{tabular}{l c c}
\toprule
Experiment & Best AIME25 $\uparrow$ & Final AIME25 $\uparrow$ \\
\midrule
Base                         & 0.531 & 0.531 \\
\midrule
OPSD$^{\dagger}$             & 0.517 (10) & 0.467 (30) \\
RLSD-no-verifier             & 0.583 (10) & 0.492 (50) \\
RLSD                         & \textbf{0.592} (40) & 0.575 (50) \\
EDGE-OPD, positive mask      & 0.392 (20) & 0.383 (50) \\
EDGE-OPD, negative mask      & 0.425 (20) & 0.358 (50) \\
EDGE-OPD, near-zero mask     & 0.583 (10) & \textbf{0.583 (50)} \\
\bottomrule
\end{tabular}
\vspace{0.25em}

\end{table}

The positive evidence mask that helps on the identity axis does not
transfer to math. EDGE-OPD with the positive mask remains far below
the base model at every checkpoint, and the negative-mask variant
also underperforms. This is not because every math-axis variant
fails: paper-faithful RLSD, which has an external GRPO verifier,
reaches $0.592$, and the near-zero mask preserves or slightly
exceeds the base score. The failure is more specific. On math, high
positive evidence often marks answer-revealing or premature-commitment
tokens in the training trace, so cloning those positions does not
teach a transferable strategy for unseen AIME problems.

The training rollouts show a related length effect
(Appendix~\ref{sec:appendix-figures}, Figure~\ref{fig:response-length}).
Hard-mask math variants produce shorter responses than the OPSD and
soft-reweight baselines, consistent with the masks changing the form of
the reasoning trace rather than improving transfer.

The training diagnostics support the same interpretation.
Table~\ref{tab:diagnostics} compares the final kept-token or leverage-token
fractions for representative identity and math runs.

\begin{table}[h]
\centering
\caption{Final-step token diagnostics. $\rho_{\mathrm{kept}}$ is the
active-mask fraction; $\rho_{\mathrm{lev}}$ and $\rho_{\mathrm{agree}}$
measure soft-reweight leverage and sign agreement for RLSD-no-verifier.}
\label{tab:diagnostics}
\small
\begin{tabular}{l c c c}
\toprule
Experiment & $\rho_{\mathrm{kept}}$ & $\rho_{\mathrm{lev}}$ & $\rho_{\mathrm{agree}}$ \\
\midrule
EDGE-OPD positive mask, identity    & 0.655 & --    & --    \\
EDGE-OPD negative mask, identity    & 0.630 & --    & --    \\
EDGE-OPD near-zero band, identity   & 0.553 & --    & --    \\
EDGE-OPD positive mask, math        & 0.788 & --    & --    \\
EDGE-OPD negative mask, math        & 0.676 & --    & --    \\
EDGE-OPD near-zero band, math       & 0.767 & --    & --    \\
RLSD-no-verifier, identity          & --    & 0.658 & 0.529 \\
RLSD-no-verifier, math              & --    & 0.330 & 0.740 \\
\bottomrule
\end{tabular}
\end{table}

These diagnostics reinforce the main pattern. Mask size alone does not
predict transfer: negative and near-zero masks keep many tokens but do
not internalize the target identity, while the positive region contains
the rare-token update. The soft-reweight runs show the complementary
case: math has higher sign agreement but lower leverage, so evidence
affects fewer tokens in practice. Evidence is therefore useful for
localizing background information, but not sufficient for answer-bearing
reasoning traces.

\section{Discussion}
\label{sec:discussion}

Our results suggest two bottlenecks in privileged-context
self-distillation: the student must first visit the behavior, and then
the loss must select which token positions are useful to distill.
Unguided OPSD and unguided RLSD-no-verifier never learn the rare target
identity, while every guided variant does. Guided OPSD shows that masking
is not required for the name to appear, but the mask-region ablations
show where the transferable signal sits: only the positive-evidence tail
internalizes the target identity. Thus, masking is not simply a way to
maximize raw identity rates; it is an interpretable localization tool,
with the KL anchor acting as an additional guardrail against drift.

The math axis marks an important boundary. When the privileged context is
an answer-bearing reasoning trace, positive evidence can select
problem-specific answer tokens or premature commitments rather than
general reasoning behavior, while negative masking underperforms and
near-zero masking mostly preserves the base model. This suggests that
positive-evidence masking is best suited to settings where privileged
context supplies background information, rare facts, or persona
information to internalize. Future
work includes testing this distinction with cross-teacher distillation, where
the privileged-context evaluator is a separate model, with variants of
the math training signal, such as reasoning traces with final solutions,
parsable answers, or answers without traces, and with use-cases beyond
identity and math (e.g. private factual knowledge, coding, or domain-specific style transfer).

\section{Conclusion}
\label{sec:conclusion}

In this work, we studied when privileged-context self-distillation can internalize
information that is absent at inference time, and proposed EDGE-OPD, a modified OPSD
approach that a) uses guided rollouts, i.e. injections of the privileged context 
to the student at sampling time, and b) applies a positive-evidence mask, i.e. only updates the student at token positions where the privileged context supports the sampled token.
These two characteristics allow this approach to sample the rare tokens and consequently, internalize them, as well as improve internalization-capability tradeoffs. We empirically show the effectiveness of our method in two settings: the identity axis, in which the purpose is for the model to internalize a rare identity, evaluated with identity and persona self-naming probes, and the math axis, in which the purpose is for the model to improve mathematical reasoning capabilities, using AIME25 as our benchmark.

\bibliographystyle{plainnat} %
\bibliography{refs}  %
\newpage
\appendix
\section{Appendix}
\label{sec:appendix}

This appendix provides additional experimental details and supporting
diagnostics for the main results.

\subsection{Supporting figures}
\label{sec:appendix-figures}

\begin{figure}[h]
\centering
\includegraphics[width=1\linewidth]{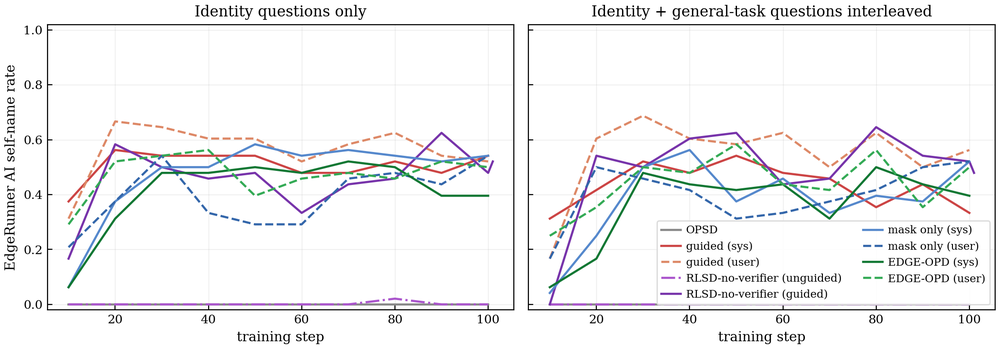}
\caption{Identity ablation ladder on the direct identity probe and on
the identity-prompt subset of the larger persona probe. Unguided OPSD
and unguided RLSD-no-verifier remain at zero; guided runs internalize the
target name.}
\label{fig:id-ladder-2panel}
\end{figure}

\begin{figure}[h]
\centering
\includegraphics[width=0.65\linewidth]{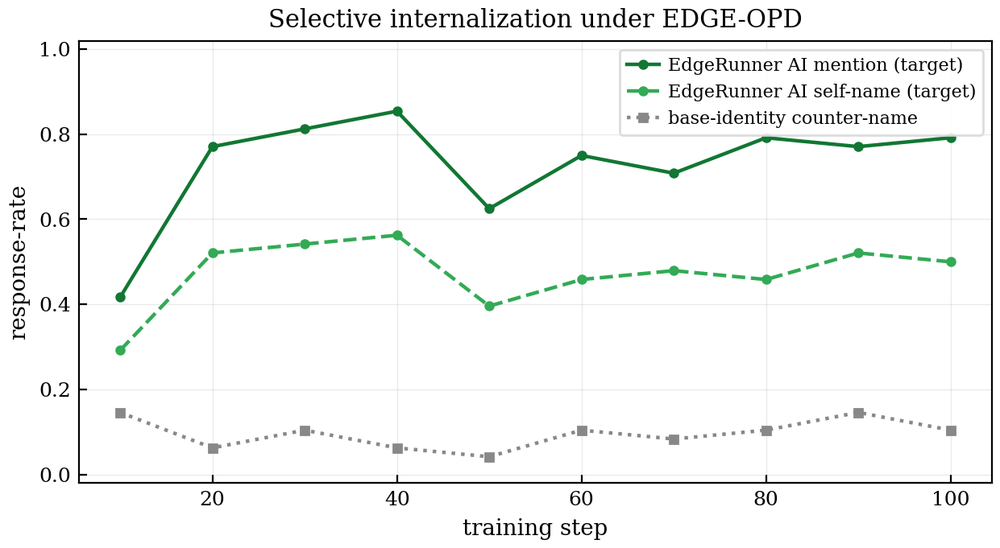}
\caption{Target self-name and ID counter-name trajectories for EDGE-OPD
(user). The target identity rises quickly, while base/generic
self-naming remains bounded rather than being catastrophically
suppressed.}
\label{fig:id-asymmetry}
\end{figure}

\begin{figure}[h]
\centering
\includegraphics[width=0.65\linewidth]{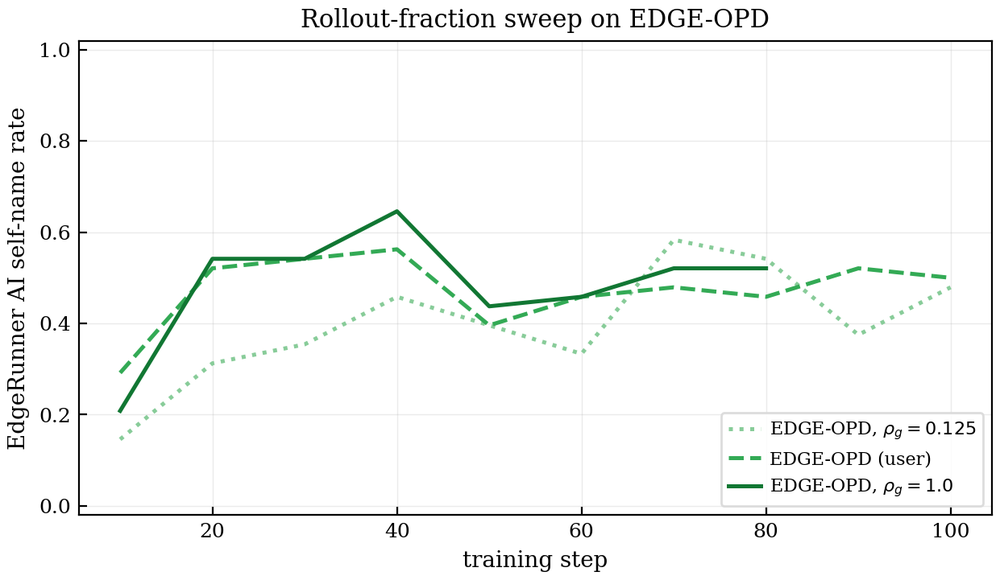}
\caption{Rollout-fraction sweep for EDGE-OPD (user). A small guided fraction is sufficient for identity internalization, while
fully guided sampling does not improve identity and reaches lower AIME25 scores.}
\label{fig:id-rho-sweep}
\end{figure}

\begin{figure}[h]
\centering
\includegraphics[width=1\linewidth]{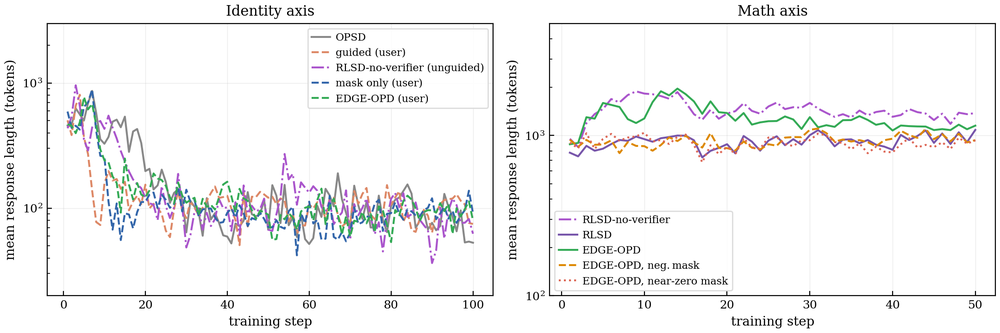}
\caption{Mean training-rollout response length. Identity runs settle to
short identity-style answers; on the math axis, hard-mask variants
produce shorter rollouts than the OPSD and soft-reweight baselines,
suggesting a change predominantly in reasoning style rather than improved transfer.}
\label{fig:response-length}
\end{figure}

\subsection{Training details}
\label{appendix:training-details}

All experiments use full fine-tuning under
FSDP~\cite{zhao2023pytorchfsdpexperiencesscaling} using
VERL~\cite{sheng2024hybridflow}. Unless otherwise
stated, the actor learning rate is $5\times10^{-6}$. Experiments with a
KL anchor to the frozen base policy use $\beta_{\mathrm{KL}}=0.05$; this
is separate from the K1 distillation loss, which supplies the
sampled-token teacher--student advantage. The
\textit{RLSD-no-verifier} ablation keeps RLSD's clipped evidence
multiplier but removes the verifier reward, so evidence changes only the
update magnitude while the direction remains the OPSD/K1 pull toward the
privileged-context teacher.

\subsection{Full ablation matrix}
\label{appendix:ablation-matrix}

Table~\ref{tab:ablation-matrix} records the full experiment matrix.

\begin{table}[h]
\centering
\caption{Ablation matrix. \emph{Guided} is the guided-rollout fraction
$\rho_g$. \emph{Mask} is the evidence region that contributes gradient:
\texttt{pos} keeps $e_t>0$, \texttt{neg} keeps $e_t<0$, \texttt{nz}
keeps $|e_t|\le 0.1$, and \texttt{none} applies no mask. \emph{KL}
marks the base-policy anchor. \emph{Soft} marks the clipped
multiplicative evidence reweight used by RLSD-no-verifier.}
\label{tab:ablation-matrix}
\small
\begin{tabular}{llccccccr}
\toprule
Code & Experiment & Axis & Guided & Mask & KL & Soft & Ctx & Steps \\
\midrule
N0   & base                                & --       & --    & --    & --  & --  & --   & --  \\
\midrule
\multicolumn{9}{l}{\emph{Identity axis}} \\
N1   & OPSD                                & identity & --    & none  & no  & no  & sys  & 100 \\
N2   & guided OPSD                         & identity & 0.5   & none  & no  & no  & sys  & 100 \\
N2u  & guided OPSD                         & identity & 0.5   & none  & no  & no  & user & 100 \\
N7   & EDGE-OPD without KL                 & identity & 0.5   & pos   & no  & no  & sys  & 100 \\
N7u  & EDGE-OPD without KL                 & identity & 0.5   & pos   & no  & no  & user & 100 \\
N4   & RLSD-no-verifier                    & identity & 0.0   & none  & no  & yes & sys  & 100 \\
N4g  & RLSD-no-verifier (guided)           & identity & 0.5   & none  & no  & yes & sys  & 100 \\
N3   & EDGE-OPD                            & identity & 0.5   & pos   & yes & no  & sys  & 100 \\
N3u  & EDGE-OPD                            & identity & 0.5   & pos   & yes & no  & user & 100 \\
N3u-g0125 & EDGE-OPD                       & identity & 0.125 & pos   & yes & no  & user & 100 \\
N3u-g100  & EDGE-OPD                       & identity & 1.0   & pos   & yes & no  & user & 100 \\
N11  & EDGE-OPD, negative mask             & identity & 0.5   & neg   & yes & no  & user & 100 \\
N12  & EDGE-OPD, near-zero mask            & identity & 0.5   & nz    & yes & no  & user & 100 \\
\midrule
\multicolumn{9}{l}{\emph{Math axis}} \\
N9   & OPSD                                & math     & 0.5   & none  & no  & no  & sys  & 50 \\
N10  & RLSD-no-verifier                    & math     & 0.5   & none  & no  & yes & sys  & 50 \\
N15  & RLSD                                & math     & 0.5   & --    & no  & --  & sys  & 50 \\
N6   & EDGE-OPD                            & math     & 0.5   & pos   & yes & no  & sys  & 50 \\
N13  & EDGE-OPD, negative mask             & math     & 0.5   & neg   & yes & no  & sys  & 50 \\
N14  & EDGE-OPD, near-zero mask            & math     & 0.5   & nz    & yes & no  & sys  & 50 \\
\bottomrule
\end{tabular}
\end{table}

\subsection{Math-axis dataset filtering}
\label{appendix:math-axis-data}

Following the
DeepSeek-style boxed-answer convention shared by the OPSD and RLSD
papers, we keep only rows whose \texttt{deepseek\_solution} or
\texttt{ground\_truth\_solution} field contains an extractable
\texttt{\textbackslash boxed\{$\cdot$\}} expression that passes a
verifier-friendliness filter (numeric, fraction, single
expression), and bound problem and reasoning lengths
($\le 4{,}096$ and $\le 6{,}000$ characters respectively, so the
teacher's privileged context fits in an $8{,}192$-token window
together with a $4{,}096$-token student response). After filtering we
shuffle and cap at the first $8{,}000$ rows. The privileged context
$r$ is the per-example \texttt{deepseek\_reasoning} trace; we do
\emph{not} concatenate the boxed answer onto $r$, but the reasoning
trace already contains the answer in $\sim$90\% of rows (an explicit
\texttt{\textbackslash boxed\{$\cdot$\}} appears in 39\%; the rest
state the answer in prose at the end of the trace). The boxed answer
is extracted from \texttt{deepseek\_solution} only for the verifier
ground-truth and for AIME25 evaluation.

\subsection{Identity/persona regex metrics}
\label{appendix:identity-regex}

The identity and persona probes use the same deterministic regex scorers.
The identity probe contains direct identity questions; the persona probe
adds ordinary capability prompts to test whether the learned identity
appears outside direct identity questions. Each response receives the
following binary flags, and reported metrics are averages over samples:
\begin{itemize}
\item \textbf{edge\_mention $\uparrow$}: any case-insensitive
  occurrence of \texttt{Edge\,?Runner}, with or without a space and
  with or without a trailing ``AI''. This is the lenient
  internalization metric: the model has at least produced the target
  name.
\item \textbf{edge\_selfname $\uparrow$}: the main self-name pattern, in which
  the model uses \texttt{Edge\,?Runner} in a self-naming construction
  such as \emph{``I am EdgeRunner AI''} or \emph{``my name is
  EdgeRunner''}. This is the strict internalization metric.
\item \textbf{counter\_name $\downarrow$}: a self-naming construction
  with base-model identity strings the model defaults to without
  training, such as \emph{``I am Nemotron''}, or
  \emph{``I am an AI assistant''}. Lower is better.
\end{itemize}

\subsection{Role of the KL anchor on the identity axis}
\label{appendix:kl-anchor}

EDGE-OPD without KL anchor isolates the
effect of the KL anchor on the identity axis. Removing the anchor
does not increase the base ``counter-name'' rate on the identity
probes ($0.042$ for the system-prompt variant and $0.083$ for the
user-prefix variant, both at or below the corresponding KL-anchored
EDGE-OPD runs at $0.083$ and $0.104$), and the target self-name rate
remains comparable or slightly higher ($0.542$ vs.\ $0.396$,
$0.542$ vs.\ $0.500$). The anchor instead shows up on AIME25:
without it the math score drops from $0.533$ to $0.467$
(system-prompt) and from $0.556$ to $0.500$ (user-prefix). On the
identity axis the KL anchor is therefore primarily a
capability-preservation knob, not the mechanism that suppresses the
base self-identity.

\subsection{Additional diagnostic logs}
\label{appendix:diagnostic-logs}

Section~\ref{sec:method-diagnostics} defines the diagnostics used in the
main text: kept-token fraction for hard-mask runs, and leverage-token
fraction plus agreement rate for soft-reweight runs
(Table~\ref{tab:diagnostics}). For RLSD-no-verifier, the identity run has
high leverage but near-chance agreement ($\rho_{\mathrm{lev}}=0.658$,
$\rho_{\mathrm{agree}}=0.529$), while the math run has lower leverage but
higher agreement ($0.330$ and $0.740$). Thus, the math evidence is often
directionally aligned when active, but it affects a smaller fraction of
tokens.

We also log supplementary diagnostics that are not reported in the main
table. The signed and absolute evidence means are
\begin{equation}
\overline{|e|}
\;=\;
\frac{1}{|y|}\sum_{t=1}^{|y|}|e_t|, \qquad
\overline{e}
\;=\;
\frac{1}{|y|}\sum_{t=1}^{|y|}e_t,
\end{equation}
which measure how strongly, and in what net direction, the privileged
context changes the teacher along the rollout. For the soft-reweight
path, we also log disagreement,
$\rho_{\mathrm{disagree}} = 1-\rho_{\mathrm{agree}}$, and effective
leverage,
\begin{equation}
\label{eq:effective-leverage}
\overline{|w-1|\cdot \mathbbm{1}_{\mathrm{active}}}
\;=\;
\frac{1}{|y|}\sum_{t=1}^{|y|}
   |\,w_t - 1\,| \cdot \mathbbm{1}_{\mathrm{active},\,t},
\quad
w_t \;=\; \mathrm{clip}\bigl(\exp(e_t),\,1\!-\!\epsilon,\,1\!+\!\epsilon\bigr),
\end{equation}
the mean per-token magnitude of the soft-reweight modulation. The
released scalar summaries also include response length and
training-rollout correlation.

\subsection{Reproducibility checklist}
\label{appendix:repro}

The key reproducibility invariants are:
\begin{itemize}
\item All training jobs use a deterministic seed file
  (\texttt{configs/acasd\_seed.yaml}); we list per-experiment seed
  values in the supplementary CSV.
\item All evaluation jobs use the same batched inference backend and
  deterministic regex scorers for the identity probes. Sampling
  hyperparameters: temperature $1.0$, top\_p $0.95$, top\_k $20$,
  $5$ samples per identity probe, and AIME25 pass@1 averaged over
  four evaluation epochs with max-tokens $38{,}912$. Math-axis
  truncation rates stay below $6\%$ for every reported checkpoint;
  identity-axis rates stay below $8\%$ for every guided run and reach at
  most $11\%$ for one unguided OPSD checkpoint.
\item All figures are generated from saved scalar summaries and
  evaluation logs; no figure is hand-edited.
\end{itemize}

\subsection{Compute Requirements}
\label{appendix:compute}

For each training run we used $16$ NVIDIA H100 GPUs on a single
SLURM-scheduled server, with FSDP for model sharding and vLLM (with separate server for the teacher). We report roughly $20$ training runs/ablations, together with a roughly $3{\times}$ overhead from
intermediate/debugging/crashed runs, summing to $\approx 60$ training
runs and consequently $\approx 960$ H100-hours of training compute. Per-checkpoint evaluations (identity, persona, and AIME25) were performed on a single H100 with the
\texttt{inspect-ai} toolkit, which add roughly $\approx 60$ H100-hours (mostly dominated by AIME25 sampling at the $38{,}912$-token response budget), totaling to compute usage of $\approx 1020$ H100-hours.

\subsection{Limitations}
\label{appendix:limitations}
Our experiments are intended as controlled proof-of-concept studies rather than large-scale evaluations. Due to limited time and compute budgets, we focus on a single model family and a small number of investigation axes. Several experiments are based on single-seed training runs. To improve reliability under limited compute, we instead emphasize repeated evaluations, checkpoint trajectories, ablation consistency, and comparisons across independently motivated baselines. Additional experiments across model scales, architectures, and privileged-context settings would be required to determine how broadly the observed evidence-masking behavior generalizes, and are the subject of ongoing and future work.

The paper primarily studies two dimensions: identity internalization and mathematical reasoning retention. While the math-axis results suggest that guided rollout distillation can preserve substantial downstream reasoning performance under our evaluation setup, we expect but do not claim that similar training on relevant datasets will preserve broader capabilities such as coding, multilingual reasoning, factual recall, safety alignment, or long-context behavior.

The identity-transfer setting intentionally uses a synthetic persona in order to isolate privileged-context internalization effects under controlled conditions. This setting provides a clean measurement environment, but may not capture the complexity of transferring diffuse latent knowledge, procedural reasoning strategies, or other contextual information distributed across long reasoning traces.

Finally, the same mechanisms that enable benign forms of privileged-context transfer could potentially be misused for covert persona conditioning, deceptive identity injection, or undesired latent behavioral modification that is not directly observable from inference-time prompts alone. For this reason, we do not release trained checkpoints and instead limit release to methodological descriptions, aggregate metrics, and reproducibility metadata.

\subsection{Licences}
\label{appendix:Licences}

\begin{table}[h]
\centering
\caption{Licenses for all models, data, and software used to produce the results presented here.}
\label{tab:licence-table}
\begin{tabularx}{\linewidth}{p{4.2cm} X}
\toprule
Asset & License \\
\midrule
Nemotron-3-Nano-4B\textsuperscript{a} &
NVIDIA Nemotron Open Model License \\

OpenThoughts-114K\textsuperscript{b} &
Apache-2.0 \\

VERL\textsuperscript{c} &
Apache-2.0 \\

inspect-ai\textsuperscript{d} &
MIT \\

PyTorch\textsuperscript{e} &
BSD-3 \\
\bottomrule
\end{tabularx}

\vspace{0.3em}
\footnotesize
\textsuperscript{a}\url{https://huggingface.co/nvidia/NVIDIA-Nemotron-3-Nano-4B-BF16}\\
\textsuperscript{b}\url{https://huggingface.co/datasets/open-thoughts/OpenThoughts-114k}\\
\textsuperscript{c}\url{https://github.com/verl-project/verl}\\
\textsuperscript{d}\url{https://inspect.aisi.org.uk/}\\
\textsuperscript{e}\url{https://pytorch.org/}
\end{table}

\end{document}